\documentclass[11pt,letterpaper]{article}
\usepackage{acl2016}
\usepackage{amsmath}
\usepackage{paralist}
\usepackage{subfig}
\usepackage{times}
\usepackage{latexsym}
\usepackage{graphicx}
\usepackage[T1]{fontenc}
\usepackage{tikz}
\usepackage{url}
\usepackage{pgfplotstable}
\usepackage{titlesec}
\usepackage{color}
\usepackage{lipsum,adjustbox}
\usepackage[font={small}]{caption}
\usetikzlibrary{positioning}

\aclfinalcopy



\newcommand{\com}[1]{}
\newcommand{\secref}[1]{\S\ref{#1}}
\newcommand{\figref}[1]{Figure~\ref{#1}}
\newcommand{\tabref}[1]{Table~\ref{#1}}
\newcommand{\oa}[1]{\footnote{\color{red}OA: #1}}

\def\perscite#1{\newcite{#1}}
\def\parcite#1{\cite{#1}}

\def\func#1{\text{\it #1}}  
\def\HUME{\func{HUME}}
\def\Adequate{\func{Adequate}}
\def\Green{\func{Green}}
\def\Orange{\func{Orange}}
\def\Units{\func{Units}}

\newcommand{\GrantNo}{644402}
\newcommand{\ProjectName}{HimL}

\title{HUME: Human UCCA-Based Evaluation of Machine Translation}

\author{
  Alexandra Birch\textsuperscript{1}\thanks{$^*$ All authors contributed equally to this work.} , Omri Abend\textsuperscript{2*}, Ond{\v{r}}ej Bojar\textsuperscript{3*},
  Barry Haddow\textsuperscript{1*}\\
\textsuperscript{1}School of Informatics, University of Edinburgh\\
\textsuperscript{2}School of Computer Science and Engineering, Hebrew University of Jerusalem\\
\textsuperscript{3} Charles University in Prague, Faculty of Mathematics and Physics\\
\texttt{a.birch@ed.ac.uk, oabend@cs.huji.ac.il}\\
\texttt{bojar@ufal.mff.cuni.cz, bhaddow@inf.ed.ac.uk}\\
}

\date{}

\begin{document}

\maketitle

\begin{abstract}
  

Human evaluation of machine translation normally uses sentence-level measures such as relative ranking or adequacy scales. However, these provide no insight into possible errors,
and do not scale well with sentence length.
We argue for a semantics-based evaluation, which captures what meaning components
are retained in the MT output, thus providing a more fine-grained analysis of
translation quality, and enabling the construction and tuning of semantics-based MT. 
We present a novel human semantic evaluation measure, Human
UCCA-based MT Evaluation (HUME), building on the UCCA semantic representation scheme.
HUME covers
a wider range of semantic phenomena than previous methods and does not rely on semantic annotation
of the potentially garbled MT output. 
We experiment with four language pairs, demonstrating HUME's broad applicability,
and report good
inter-annotator agreement rates and 
 correlation with human adequacy scores.


\end{abstract}

\section{Introduction}\label{sec:intro}


Human judgement should be the ultimate test of the quality of an MT system.
Nevertheless, common measures for human MT evaluation, such as adequacy and fluency judgements
or the relative ranking of possible translations, are problematic in two ways.
First, as the quality of translation is multi-faceted, it is difficult
to quantify the quality of the entire sentence in a single number. This
is indeed reflected in the diminishing inter-annotator agreement (IAA) rates of human ranking measures
with the sentence length \cite{Bojar:2011}.
Second, a sentence-level quality score does not indicate what parts of the sentence
are badly translated, and so cannot inform developers in repairing these errors.

These problems are partially addressed by measures that decompose over parts of the evaluated
translation, often words or n-grams (see \secref{sec:background} for a brief survey of previous
work). A promising line of research decomposes metrics
over semantically defined units, quantifying the similarity of the output and the
reference in terms of their verb argument structure; the most notable of these measures is HMEANT
\parcite{lo2011structured}.



We propose the HUME metric,
a human evaluation measure that decomposes over UCCA semantic units.
UCCA \parcite{abend2013universal} is an appealing candidate for semantic analysis,
due to its cross-linguistic applicability, support for rapid annotation, and coverage
of many fundamental semantic phenomena, such as verbal, nominal and adjectival
argument structures and their inter-relations.

HUME operates by aggregating human assessments of the translation quality of individual
semantic units in the source sentence. We 
are thus avoiding the semantic annotation of machine-generated text,
which is often garbled or semantically unclear.
This also allows the re-use of the source semantic annotation for
measuring the quality of different translations of the same source sentence
and avoids relying on reference translations, 
which have been shown to bias annotators~\cite{fomicheva-specia_ACL:2016}.


After a brief review (\secref{sec:background}), we describe HUME in detail
(\secref{sec:hume}). 
Our experiments with four language pairs: English to Czech, German, Polish and Romanian (\secref{sec:experiments}) document HUME's inter-annotator agreement and efficiency (time of annotation). We further empirically compare HUME with direct assessment of human adequacy ratings (\secref{sec:adequacy}), and conclude by discussing the differences with HMEANT (\secref{sec:hmeant_comp}).

%

\section{Background}\label{sec:background}



\paragraph{MT Evaluation.}
Human evaluation is generally done by ranking the outputs of multiple systems
e.g., in the WMT tasks \parcite{bojar2015findings}, or by assigning
adequacy/fluency scores to each translation, a procedure recently improved
by \perscite{graham2015accurate} under the title Direct Assessment. We use this latter method to compare and
contrast
with HUME later in the paper. HTER~\parcite{snover2006study} is another widely used human evaluation metric
which uses edit distance metrics to compare a translation and its human 
post-edition. HTER suffers from the problem that small edits in the translation
could in fact be serious flaws in accuracy, e.g., deleting a negation. 
Some manual measures ask annotators to explicitly mark errors, but this has
been found to have even lower agreement than ranking \parcite{lommel:etal:mqm-iaa:2014}.

However, while providing the gold standard for MT evaluation, human evaluation
is not a scalable solution.
Scalability is addressed by employing automatic and semi-automatic approximations of human
judgements. Commonly, such scores decompose over the sub-parts of
the translation, and quantify how many of these sub-parts appear in a manually
created reference translation.
This decomposition allows system developers to localize the errors.
The most commonly used measures decompose over n-grams or individual words, e.g., 
BLEU \parcite{Papineni:2002}, NIST \parcite{Doddington:2002} and METEOR \parcite{Banerjee:2005}.
Another common approach is to determine the similarity between the reference and translation
in terms of string edits \parcite{snover2006study}.
While these measures stimulated much progress in MT research by allowing
the evaluation of massive-scale experiments, the focus on words and n-grams does
not provide a good estimate of semantic correctness, and may favour shallow string-based
MT models.



In order to address this shortcoming, more recent work quantified
the similarity of the reference and translation in terms
of their structure. \perscite{liu2005syntactic} took a syntactic approach, 
using dependency grammar, and
\perscite{owczarzak2007evaluating} took a similar approach using Lexical Functional Grammar structures.
\perscite{gimenez2007linguistic} proposed to combine multiple types of information,
capturing the overlap between the translation and reference in terms of their
semantic (predicate-argument structures), lexical and morphosyntactic features.
\perscite{machacek:bojar:segranks:2015} divided the source sentences into shorter segments,
defined using a phrase structure parse, and applied human ranking to the resulting segments.

Perhaps the most notable attempt at semantic MT evaluation is MEANT and
its human variant HMEANT \parcite{lo2011structured}, which quantifies the similarity between
the reference and translation in terms of the overlap in
their verbal argument structures and associated semantic roles.
We discuss the differences between HMEANT and HUME in \secref{sec:hmeant_comp}.

%

\begin{figure}
    \begin{center}
    \includegraphics[width=0.35\textwidth]{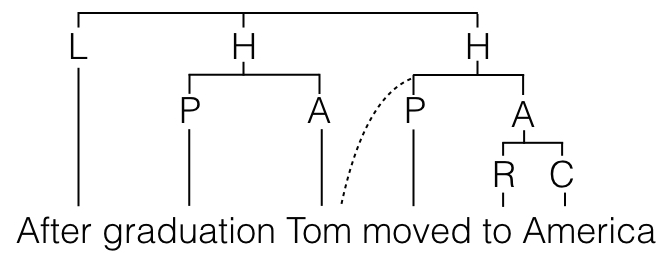}

{\small
  \begin{tabular}{|l|l|l|l|}
\hline
L & Linker &  A & Participant \\
H & Parallel Scene & R & Relater \\
P & Process & C & Centre \\
\hline
\end{tabular}
}
    \end{center}
\caption{\label{fig:ucca_example_v2}
  Sample UCCA annotation. Leaves correspond to words and nodes to units.
  The dashed edge indicates that ``Tom'' is also a participant in the ``moved to America''
  Scene. Edge labels mark UCCA categories.}
\end{figure}

\paragraph{Semantic Representation.}
UCCA (Universal Conceptual Cognitive Annotation) \cite{abend2013universal} is a
cross-linguistically applicable scheme for semantic annotation.
Formally, an UCCA structure is a directed acyclic graph (DAG),
whose leaves correspond to the words of the text.
The graph's nodes, called {\it units}, are either terminals or several elements jointly
viewed as a single entity according to some semantic or cognitive consideration. Edges bear
a category, indicating the role of the sub-unit in the structure the unit
represents.

UCCA's basic inventory of distinctions (its {\it foundational layer})
focuses on argument structures (adjectival, nominal, verbal and others)
and relations between them. The most basic notion is the {\it Scene}, which describes a movement, an
action or a state which persists in time. Each Scene contains one main
relation and zero or more participants. For example, the sentence ``After graduation, Tom moved to America''
contains two Scenes, whose main relations are ``graduation'' and ``moved''.
The participant ``Tom'' is a part of both Scenes, while ``America'' only of the
latter (\figref{fig:ucca_example_v2}). Further categories account for
inter-scene relations and the sub-structures of participants and relations.

The use of UCCA for semantic MT evaluation has several motivations. 
First, UCCA's foundational layer can be annotated by non-experts after a short training
\parcite{abend2013universal,marinotti2014}.
Second, UCCA is cross-linguistically applicable, seeking to
represent what is shared between languages by building on
linguistic typological theory \parcite{Dixon:10a,Dixon:10b,Dixon:12}.
Its cross-linguistic applicability has so far been tested in annotations of
English, French, German and Czech.
Third, the scheme has been shown to be stable across translations:
UCCA annotations of translated text usually contain the same set of relations
\parcite{sulem2015conceptual}, indicating that UCCA reflects a layer of
representation that in a correct translation is mostly shared between
the translation and the source.



The Abstract Meaning Representation (AMR) \parcite{banarescu2013abstract}
shares UCCA's motivation for defining a more complete semantic annotation.
However, using AMR is not optimal for defining a decomposition of a sentence into semantic
units as it does not anchor its semantic symbols in the text,
and thus does not provide clear decomposition of the sentence into sub-spans.
Also, AMR is more fine-grained than UCCA and consequently harder to annotate.
Other approaches represent semantic structures as bi-lexical dependencies
\parcite{sgallhp:1986,hajic2012announcing,oepen2006discriminant},
which are indeed anchored in the text, but are less suitable for MT evaluation
as they require linguistic expertise for their annotation.





\section{The HUME Measure}\label{sec:hume}

\subsection{Annotation Procedure}\label{sec:guidelines}

This section summarises the manual annotation procedure used
to compute the HUME measure. 
We denote the source sentence as $s$ and the translation as $t$. 
The procedure involves two manual steps: (1) UCCA-annotating $s$, 
(2) HUME-annotation: human judgements as to the translation quality of each semantic
unit of $s$ relative to $t$,
where units are defined according to the UCCA annotation.
UCCA annotation is performed once for every source sentence,
irrespective of the number of its translations we wish to evaluate,
and requires proficiency in the source language only.
HUME annotation requires the employment of bilingual
annotators.\footnote{Where bilingual annotators are not available,
  the evaluation could be based on the UCCA structure for the
  \emph{reference} translation. See discussion in \secref{sec:hmeant_comp}.}

\paragraph{UCCA Annotation.}
We begin by creating UCCA annotations for the source sentence, following the
UCCA guidelines.\footnote{All UCCA-related resources can be found
  here: \url{http://www.cs.huji.ac.il/~oabend/ucca.html}}
A UCCA annotation for a sentence $s$ is a labeled DAG $G$, whose leaves
are the words of $s$.
For every node in $G$, we define its {\it yield} to be its leaf descendants.
The semantic units for $s$ according to $G$ are the yields of nodes in $G$.

\paragraph{Translation Evaluation.}
HUME annotation is done by traversing the semantic units
of the source sentence, which correspond to the arguments and relations expressed
in the text, and marking the extent to which they have been correctly translated.
HUME aggregates the judgements of the users into a composite score, 
which reflects the overall extent to which the semantic content of $s$ is preserved in $t$.

Annotation of the semantic units requires first deciding whether
a unit is {\it structural}, i.e., has meaning-bearing
sub-units in the target language, or {\it atomic}.
In most cases, atomic units
correspond to individual words, but they may also correspond to
multi-word expressions that translate as one unit. For instance,
the expression ``took a shower'' is translated into the German ``duschte'',
while its individual words do not correspond to any sub-part of the German translation,
motivating the labeling the entire expression as an atomic node.
When a multi-word unit is labeled as atomic, its sub-units' annotations are ignored
in the evaluation.

Atomic units can be labelled as ``Green'' (G, correct), ``Orange'' (O, partially correct)
and ``Red'' (R, incorrect). 
Green means that the meaning of the word or phrase has been largely preserved.
Orange means that the essential meaning of the unit has been preserved,
but some part of the translation is wrong.
This is often be due to the translated word having the wrong inflection,
in a way that impacts little on the understandability of the sentence.
Red means that the essential meaning of the unit has not been captured.

Structural units have sub-units (children in the UCCA graph),
which are themselves atomic or structural.
Structural units are labeled as ``Adequate'' (A) or ``Bad'' (B), meaning
that the relation between the sub-units went wrong\footnote{
  Three labels are used with atomic units, as opposed to two labels with structural units,
as atomic units are more susceptible to slight errors.}.
We will use the example ``man bites dog'' to illustrate typical examples of why a structural node
should be labelled as ``Bad'':
incorrect ordering (``dog bites man''), 
deletion (``man bites'') and insertion (``man bites biscuit dog''). 

HUME labels reflect adequacy, rather than fluency judgements.
Specifically, annotators are instructed to
label a unit as Adequate if its translation is understandable and preserves
the meaning of the source unit, even if its fluency is impaired.

\begin{figure}
    \begin{center}
    \includegraphics[width=0.35\textwidth]{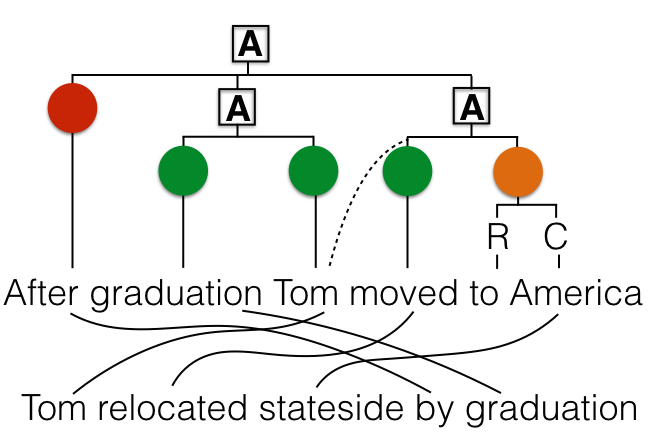}
    \end{center}
  \caption{\label{fig:hume_tree_v2}
     HUME annotation of an UCCA tree with a word-aligned example translation shown below. 
     Atomic units are labelled using traffic lights (Red, Orange, Green) and structural
     units are marked A or B.}
\end{figure}

Figure~\ref{fig:hume_tree_v2} presents an example of a HUME
annotation, where the translation is in English for ease of comprehension.
When evaluating ``to America'' the annotator looks at the translation and sees the
word ``stateside''. This word captures the whole phrase and so we mark this
non-leaf node with an atomic label. Here we choose Orange since
it approximately captures the meaning in this context.
The ability to mark non-leaves with atomic labels allows
the annotator to account for translations which only correspond at the phrase
level. Another feature highlighted in this example is that by separating structural
and atomic units, we are able to define where an error occurs, and localise
the error to its point of origin. The linker ``After'' is translated incorrectly as ``by''
which changes the meaning of the entire sentence. This error is captured at
the atomic level, and it is labelled Red. The sentence still contains two Scenes and
a Linker and therefore we mark the root node as structurally correct, Adequate.

\begin{figure*}[t]
    \begin{center}
    \includegraphics[width=.8\textwidth]{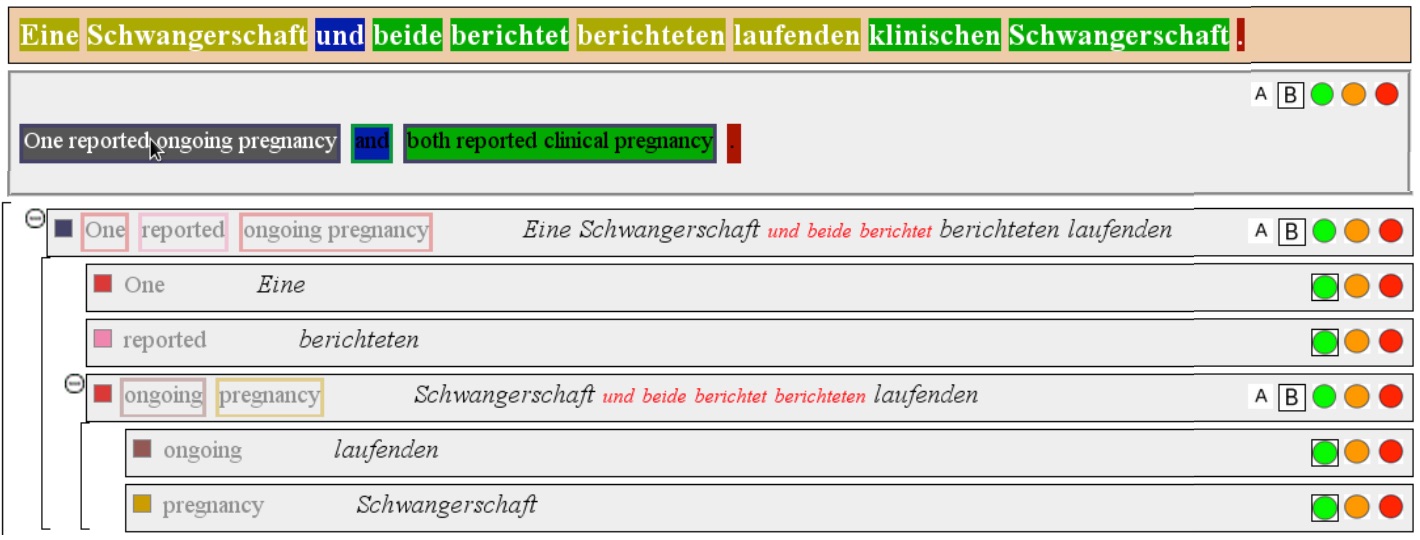}
    \caption{The HUME annotation tool. The top orange box
      contains the translation. The source sentence is directly below it, followed by the tree of the source
      semantic units. Alignments between the source and translation are in italics and
      unaligned intervening words are in red (see text).}
    \label{fig:interface}
    \end{center}
\end{figure*}

\subsection{Composite Score}\label{sec:score}

We proceed to detailing how judgements on the semantic units
of the source are aggregated into a composite score. 
We start by taking a very simple approach and compute an accuracy score.
Let $\Green(s,t)$, $\Adequate(s,t)$ and $\Orange(s,t)$ be the number of Green, Adequate and Orange
units, respectively. Let $\Units(s)$ be the number of units marked with any of the labels.
Then HUME's composite score is:

\vspace{-.3cm}

{\scriptsize
\begin{equation}
  \HUME(s,t) = \frac{\Green(s,t) + \Adequate(s,t) + 0.5\cdot \Orange(s,t)}{\Units(s)}
  \nonumber
\end{equation}
}

\vspace{-.5cm}

\subsection{Annotation Interface}

\figref{fig:interface} shows the HUME annotation interface\footnote{A demo of HUME
  can be found in \url{www.cs.huji.ac.il/~oabend/hume_demo.html}}.
One source sentence and one translation are presented at a time.
The user is asked to select a label for each source semantic unit,
by clicking the ``A'', ``B'', Green, Orange, or Red buttons to the right of the unit's box.
Units with multiple parents (as with ``Tom'' in \figref{fig:hume_tree_v2}) are displayed
twice, once under each of their parents, but are only annotatable in one of
their instances, to avoid double counting.

The interface presents, for each unit, the translation segment aligned with it.
This allows the user, especially in long sentences, to focus her attention on the parts
that are most likely to be relevant for her judgement. As the alignments are automatically derived,
and therefore noisy, the annotator is instructed to treat the aligned text is a cue, but to ignore
the alignment if it is misleading, and instead 
make a judgement according to the full translation.
Concretely, let $s$ be a source sentence, $t$ a translation,
and $A \subset 2^s \times 2^t$ a many-to-many word alignment.
If $u$ is a semantic unit in $s$, whose yield is $yld(u)$, we define the aligned text in
$t$ to be $\bigcup_{(x_s,x_t) \in A \wedge x_s \cap yld(u) \neq \emptyset} x_t$.

Where the aligned text is discontinuous in $t$, words between the left
and right boundaries which are not contained in it (intervening words)
are presented in a smaller red font. 
Intervening words are likely to change the meaning of the translation
of $u$, and thus should be attended to when considering whether the translation
is correct or not. 

For example, 
in \figref{fig:interface}, ``ongoing pregnancy'' is translated to
``Schwangerschaft ... laufenden'' (lit. ``pregnancy ... ongoing''). This alone
seems acceptable but the interleaving words in red notify the annotator to check
the whole translation, in which the meaning of the expression is not
preserved\footnote{The interleaving words are ``... und beide berichtet berichteten ...'' (lit.
  ``... and both report reported ...''), which doesn't form any coherent relation with the rest
  of the sentence.}.
The annotator should thus mark this structural node as Bad.

\section{Experiments}\label{sec:experiments}

In order to validate the HUME metric, we ran an annotation experiment with one source language (English),
and four target languages (Czech, German, Polish and Romanian), using text from the public health domain.
Semantically accurate translation is paramount in this domain, which makes
it particularly suitable for semantic  MT evaluation. HUME is evaluated in terms of its
consistency (inter-annotator agreement), efficiency (time of annotation) and validity (by comparing it
with crowd-sourced adequacy judgements).




\subsection{Datasets and Translation Systems}

For each of the four language pairs under consideration  we built phrase-based MT systems
using Moses \parcite{Koehn:2007}.  These were trained on large parallel data sets extracted from
OPUS \parcite{tiedemann:2009}, and the data sets released for the WMT14
medical translation task \parcite{bojar-EtAl:2014:W14-33}, 
giving between 45 and 85 million sentences of training data, depending on the language pair.
These translation systems were used to translate texts derived from both NHS
24\footnote{\url{http://www.nhs24.com/}} and
Cochrane\footnote{\url{http://www.cochrane.org/}} into the four languages.
NHS~24 is a public body providing healthcare and health-service
related information in Scotland; Cochrane is an international NGO
which provides independent systematic reviews on health-related research.
NHS~24 texts come from the ``Health A-Z'' section in the NHS Inform
website, and Cochrane texts come from their plain language summaries
and abstracts.


\subsection{HUME Annotation Statistics}
\label{sec:annot_stats}

The source sentences are all in English, and their UCCA annotation was performed by four
computational linguists and one linguist.
For the annotation of the MT output, we recruited two annotators for each of German, Romanian
and Polish and one main annotator for Czech. For computing Czech IAA,
several further annotators worked on a small number of 
sentences each. We treat these further annotators as one annotator, resulting in two annotators
for each language pair.
The annotators were all native speakers of the respective target languages and fluent in English.
They completed a three hour on-line training session which included a description of
UCCA and the HUME task, followed by walking through a few examples.

\tabref{tab:annot}
shows the total number of sentences and units annotated by each annotator.
Not all units in all sentences were annotated, often due to
the annotator 
accidentally missing a node.
\begin{table}
\begin{center}
{\small
\begin{tabular}{ll|cccc}
& & cs & de & pl & ro \\
\hline
\#Sentences &  Annot. 1 & 324   & 339  & 351  & 230  \\
 & Annot. 2 & 205 & 104  & 340  & 337 \\
\hline
\#Units & Annot. 1 & 8794  & 9253 & 9557  & 6152 \\
 &Annot. 2 & 5553 & 2906  & 9303  & 9228  \\
\end{tabular}
\caption{HUME-annotated \#sentences and \#units.}
\label{tab:annot}
}
\end{center}
\end{table}

\paragraph{Efficiency.}
We estimate the annotation time using the timestamps
provided by the annotation tool, which are recorded whenever an annotated sentence is
submitted. Annotators are not able to re-open a sentence once submitted. 
To estimate the annotation time, we compute the time difference between successive 
sentences, and discard outlying times, assuming annotation was not continuous in these cases.
From inspection of histograms of annotation times, we set the upper threshold at 500 seconds.
Median annotation times are presented in Table~\ref{tab:annot_times},
indicating that the annotation
of a sentence takes around 2--4 minutes, with some variation between annotators.


\begin{table}[t]
\begin{center}
{\small
\begin{tabular}{l|cccc}
& cs & de & pl & ro \\
\hline
Annot. 1 & 255 & 140  & 138 & 96 \\
Annot. 2 & $^*$ & 162 & 229 & 207 \\
\end{tabular}
\caption{Median annotation times per sentence, in seconds.
  $^*$: no timing information is available, as
  this was a collection of annotators, working in parallel.}
\label{tab:annot_times}
}
\end{center}
\end{table}


\paragraph{Inter-Annotator Agreement.}
\label{sec:iaa}

\begin{table}[t]
\begin{center}
{\small
\begin{tabular}{l|cccc}
 & cs & de & pl & ro \\
\hline
Sentences & 181 & 102 & 334 & 217 \\
\hline
All units & 4686   & 2793   & 8384   & 5604  \\
Kappa & 0.64   & 0.61   & 0.58   & 0.69  \\
\hline
Atomic units & 2982 & 1724 & 5386 & 3570 \\
Kappa & 0.54 & 0.29 & 0.54 & 0.50 \\
\hline
Structural units & 1602 & 1040 & 2655 & 1989 \\
Kappa & 0.31 & 0.44 & 0.33 & 0.58 \\
\end{tabular}
\caption{IAA for the multiply-annotated units,
measured by Cohen's Kappa. }
\label{tab:iaa}
}
\end{center}
\end{table}


In order to assess the consistency of the annotation, we measure the Inter-Annotator
Agreement (IAA) using Cohen's Kappa on the multiply-annotated units.
\tabref{tab:iaa} reports the number of units which have two annotations from
different annotators and the corresponding Kappas.
%
We report the overall Kappa, as well as separate Kappas on atomic
units (annotated as Red, Orange or Green) and structural units (annotated
as Adequate or Bad).
As expected and confirmed by confusion matrices in \figref{fig:heatmap}, there
is generally little confusion between the two types of units.
This results in the Kappa for all units being considerably higher than the Kappa
over the atomic units or structural units, where there is more internal confusion. 


\def\iaafig #1{\includegraphics[width=4cm]{iaa_heatmap_#1.png}}

\begin{figure}[t]
\renewcommand{\tabcolsep}{0pt}
\begin{tabular}{cc}

\subfloat[English-Czech]{
  \iaafig{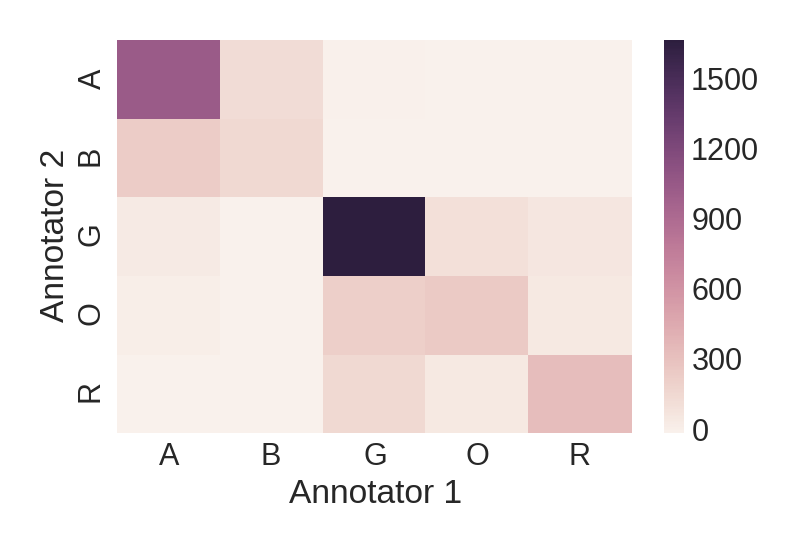}
}
&
\subfloat[English-German]{
  \iaafig{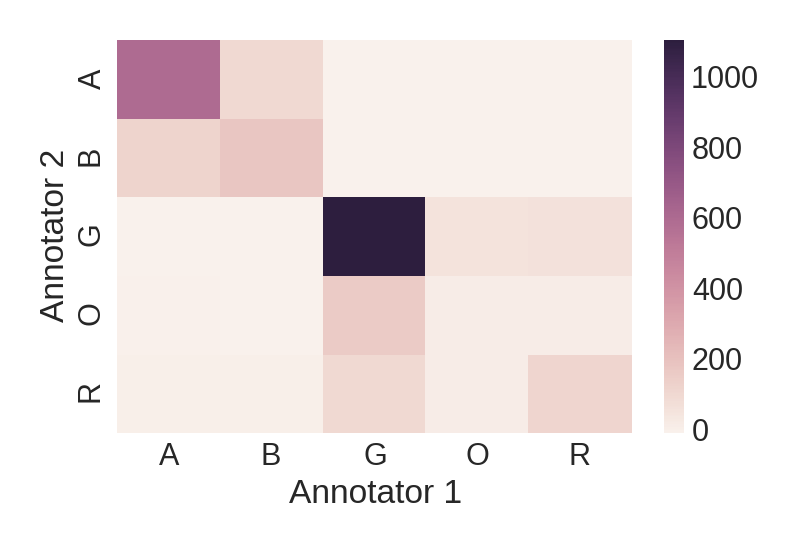}

}
\\

\subfloat[English-Polish]{
  \iaafig{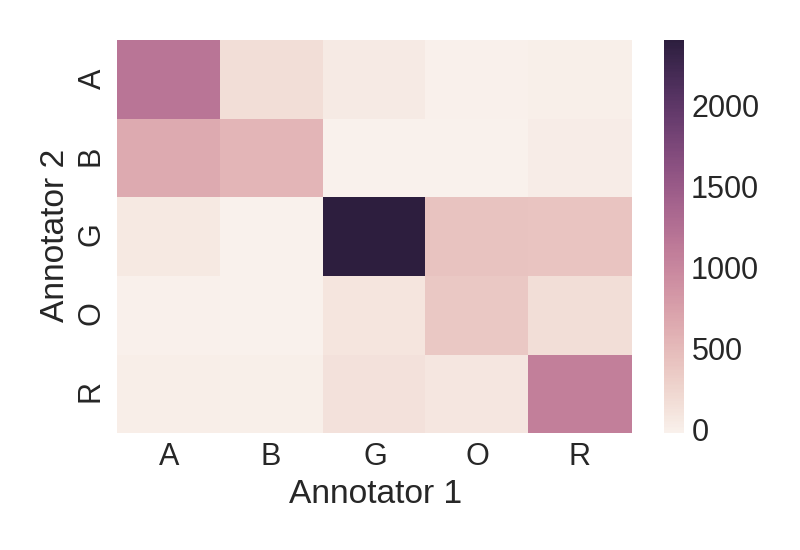}
  
}
&
\subfloat[English-Romanian]{
  \iaafig{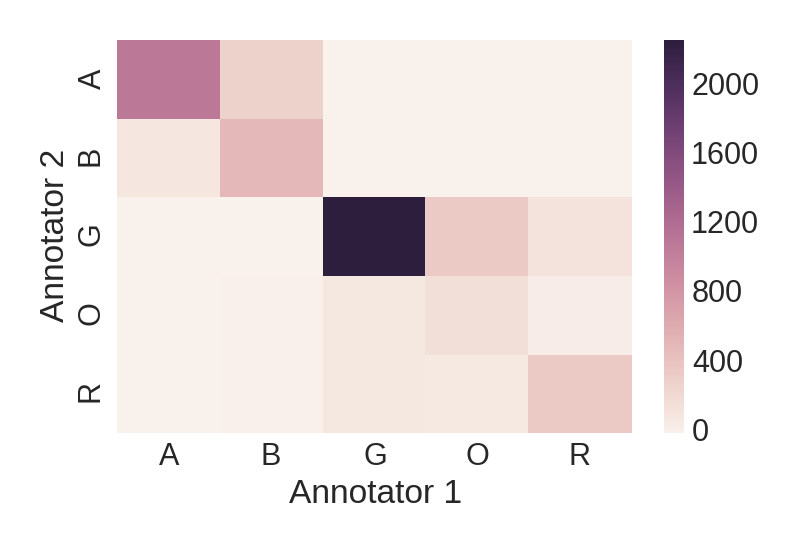}

}
\end{tabular}
\caption{Confusion matrices for each language pair.}
\label{fig:heatmap}
\end{figure}


\def\iaafig #1{\includegraphics[width=3.8cm]{iaa_length_#1.png}}

\begin{figure}[t]
\renewcommand{\tabcolsep}{0pt}
\begin{tabular}{cc}

\subfloat[English-Czech]{
  \iaafig{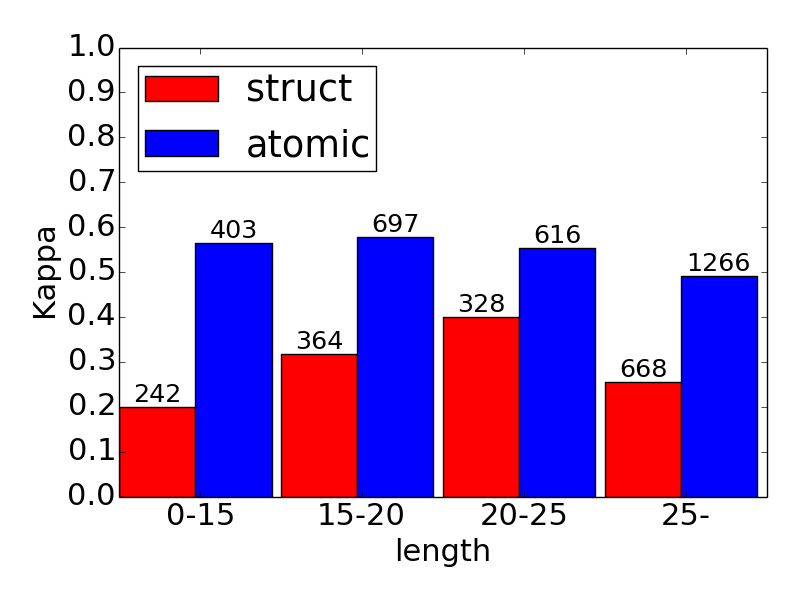}
}
&
\subfloat[English-German]{
  \iaafig{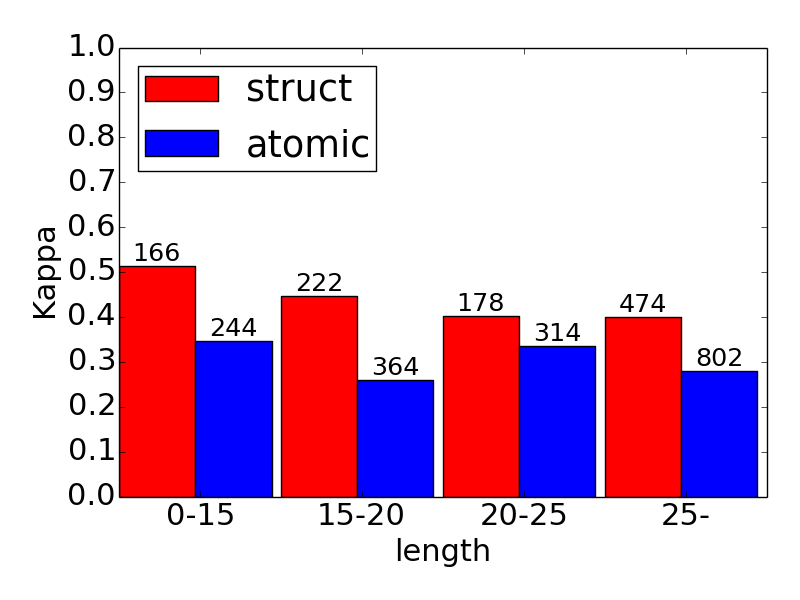}

}
\\

\subfloat[English-Polish]{
  \iaafig{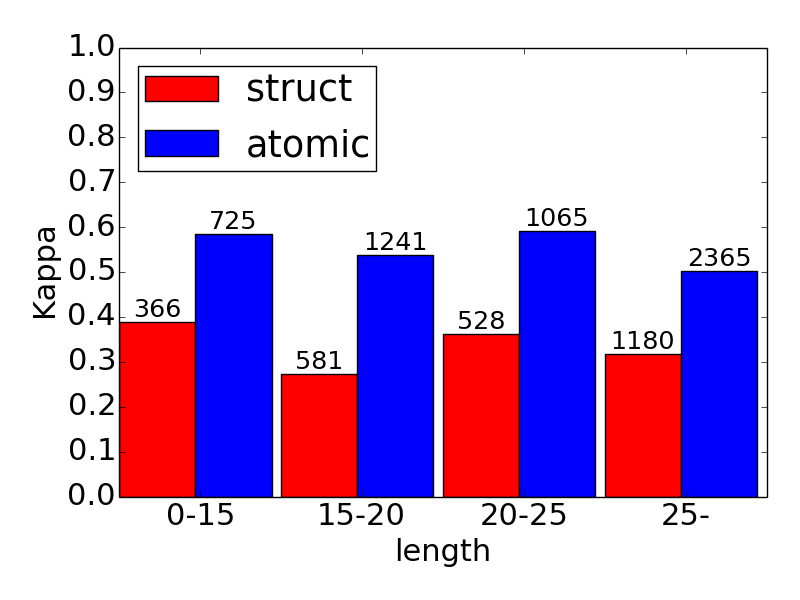}
  
}
&
\subfloat[English-Romanian]{
  \iaafig{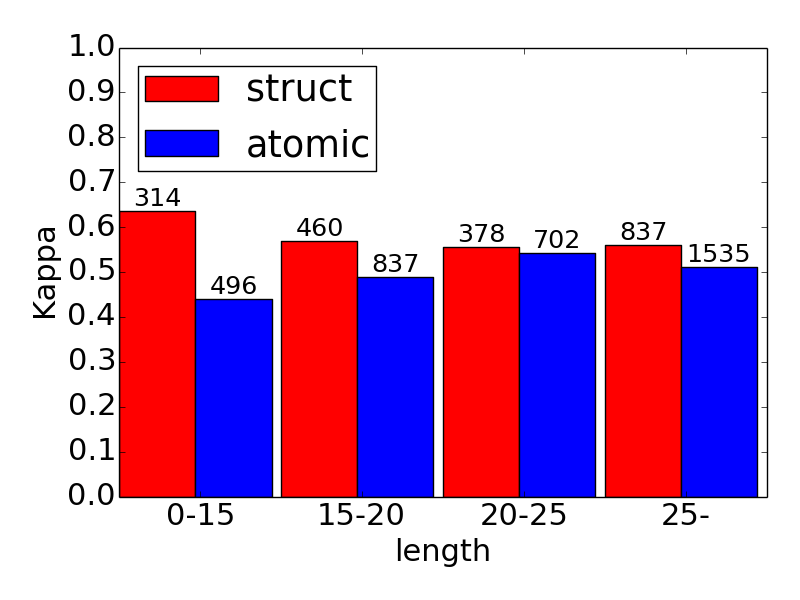}

}
\end{tabular}
\caption{Kappa versus sentence length for
structural and atomic units. (Node counts in bins on top of each bar.)
}
\label{fig:iaalength}
\end{figure}

To assess HUME reliability for long sentences,
we binned the sentences according to length and measured Kappa on each bin
(\figref{fig:iaalength}).
We see
no discernible reduction of IAA with sentence
length. \tabref{tab:iaa} also shows that the overall IAA
is similar for all languages, presenting good agreement (0.6--0.7).
However, there are differences observed when we break down by node type.
Specifically, we see a contrast  between
Czech and Polish, where the IAA is higher for atomic than for structural units, and German and Romanian,
where the reverse is true. We also observe low IAA (around 0.3) in the cases of
German atomic units, and Polish and Czech structural units.

Looking more closely at the areas of disagreement, we see that for the Polish structural units, the 
proportion of As was quite different between the two annotators (53\% vs. 71\%), whereas for other
languages the annotators agree in the proportions. We believe that this was because one of the Polish
annotators did not fully understand the guidelines for structural units, and percolated
errors up the tree, creating more Bs. For German atomic and Czech structural units, where Kappa is also around 0.3, the proportion of such units being marked as ``correct'' is relatively 
high, meaning that the class distribution is more skewed, so the expected agreement used in the
Kappa calculation is high, lowering Kappa.
Finally we note some evidence of domain-specific disagreements, for instance
the German MT system normally translated ``review'' (as in ``systematic review'' -- a frequent term in the 
Cochrane texts) as ``\"Uberpr\"ufung'', which 
one annotator marked correct, and the other (a Cochrane employee) as incorrect.

\section{Comparison with Direct Assessment}\label{sec:adequacy}

\def\iaafig #1{\includegraphics[width=0.45\textwidth]{humevsDA_10en-#1.pdf}}

\begin{figure}[t]
\renewcommand{\tabcolsep}{0pt}
\begin{tabular}{c}
\subfloat[English-German]{
  \iaafig{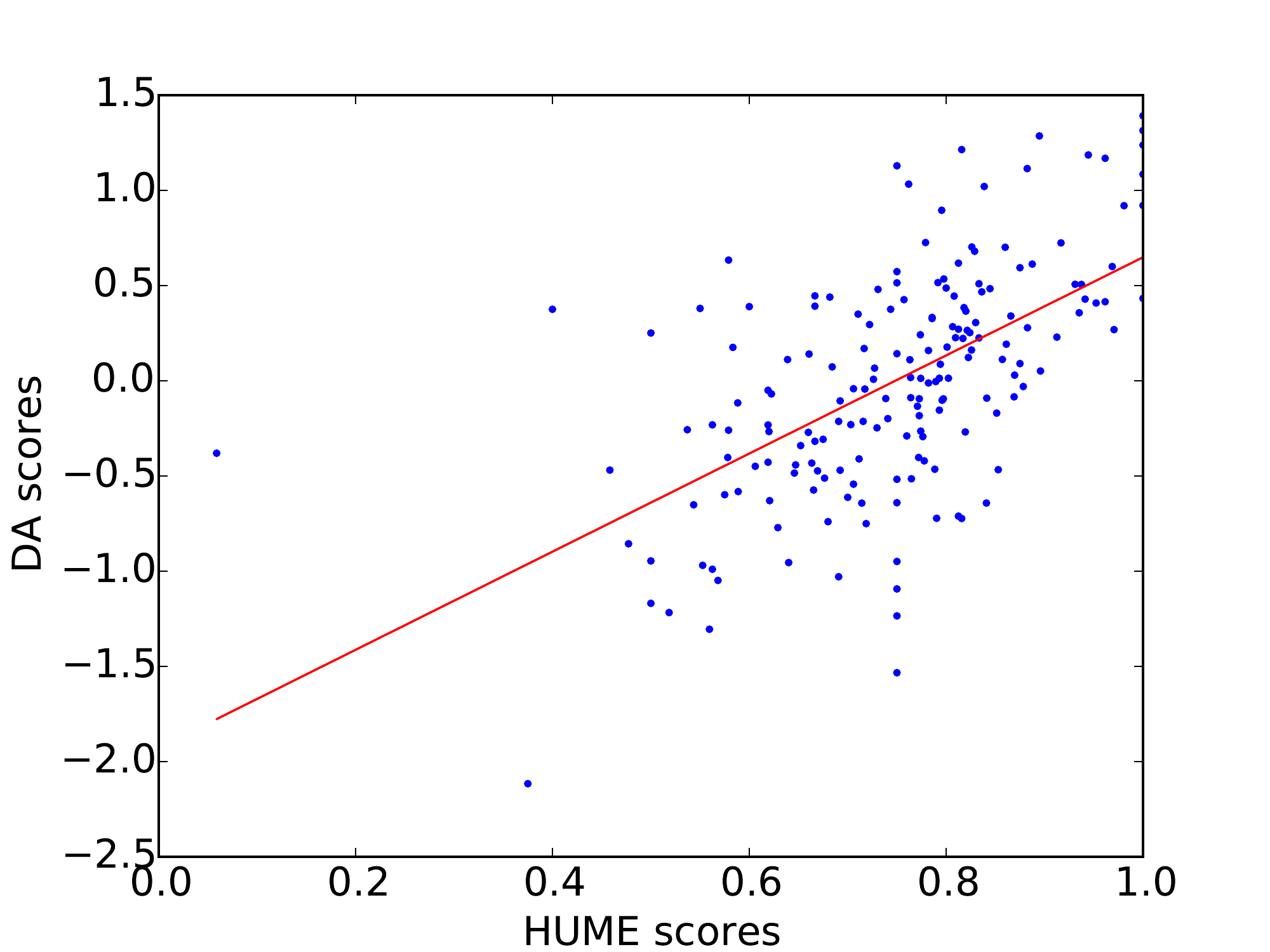}
}
\\
\subfloat[English-Romanian]{
  \iaafig{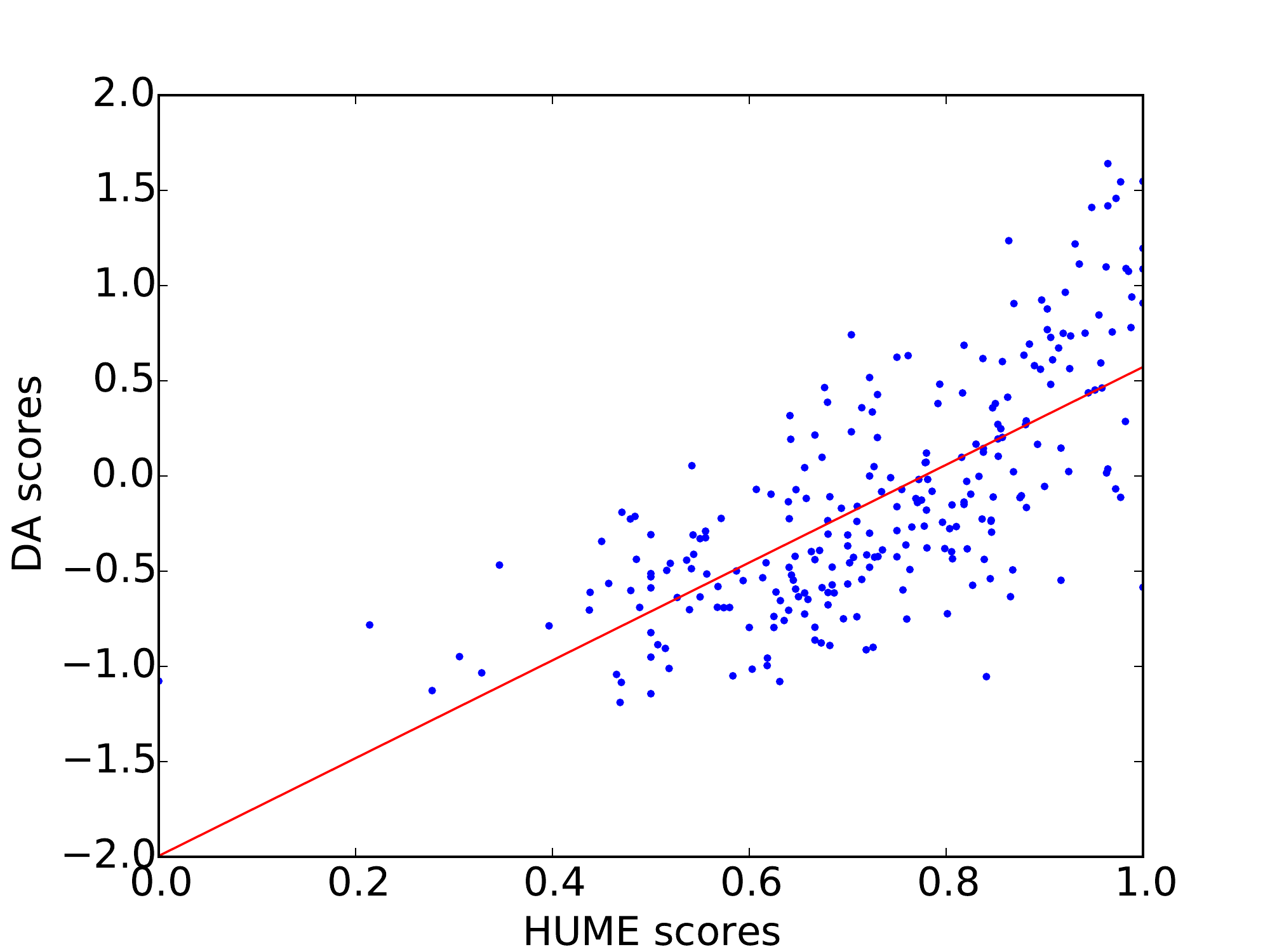}
}
\end{tabular}
\caption{HUME vs DA scores. DA score have been standardised for each crowdsourcing annotator and averaged across exactly 10 annotators. HUME scores are averaged where there were two annotations. 
}
\label{fig:dacorrelation}
\end{figure}

Recent research~\cite{graham2015accurate,graham2015crowd,graham2015improving} has proposed a new approach for collecting accuracy ratings, direct assessment (DA).
Statistical interpretation of a large number of crowd-sourced adequacy
judgements for each candidate translation on a fine-grained scale of 0 to 100
results in reliable aggregate scores, that correlate very strongly with one
another.

We attempted to follow \perscite{graham2015accurate} but struggled to get enough
crowd-sourced judgements for our target languages. We ended up with 10 adequacy 
judgements on most of the HUME annotated translations for
German and Romanian but insufficient data for Czech and Polish. We see this as a
severe practical limitation of DA.

Figure~\ref{fig:dacorrelation} plots the HUME score for each sentence against
its DA score. HUME and Direct Assessment scores correlate reasonably well. The
Pearson correlation for en-ro (en-de) is 0.70 (0.58), or 0.78 (0.74) if only
doubly HUME-annotated points are considered.
This confirms that HUME is consistent with an accepted human evaluation method,
despite their conceptual differences.
While DA is a valuable tool, HUME has two advantages:
it returns fine-grained semantic information about 
the quality of translations and it only requires very few annotators.
Direct assessment returns a single opaque score, and (as also noted by
Graham et al.) requires a large crowd which may not be available or reliable. 

\begin{figure}[t]
\includegraphics[width=0.45\textwidth]{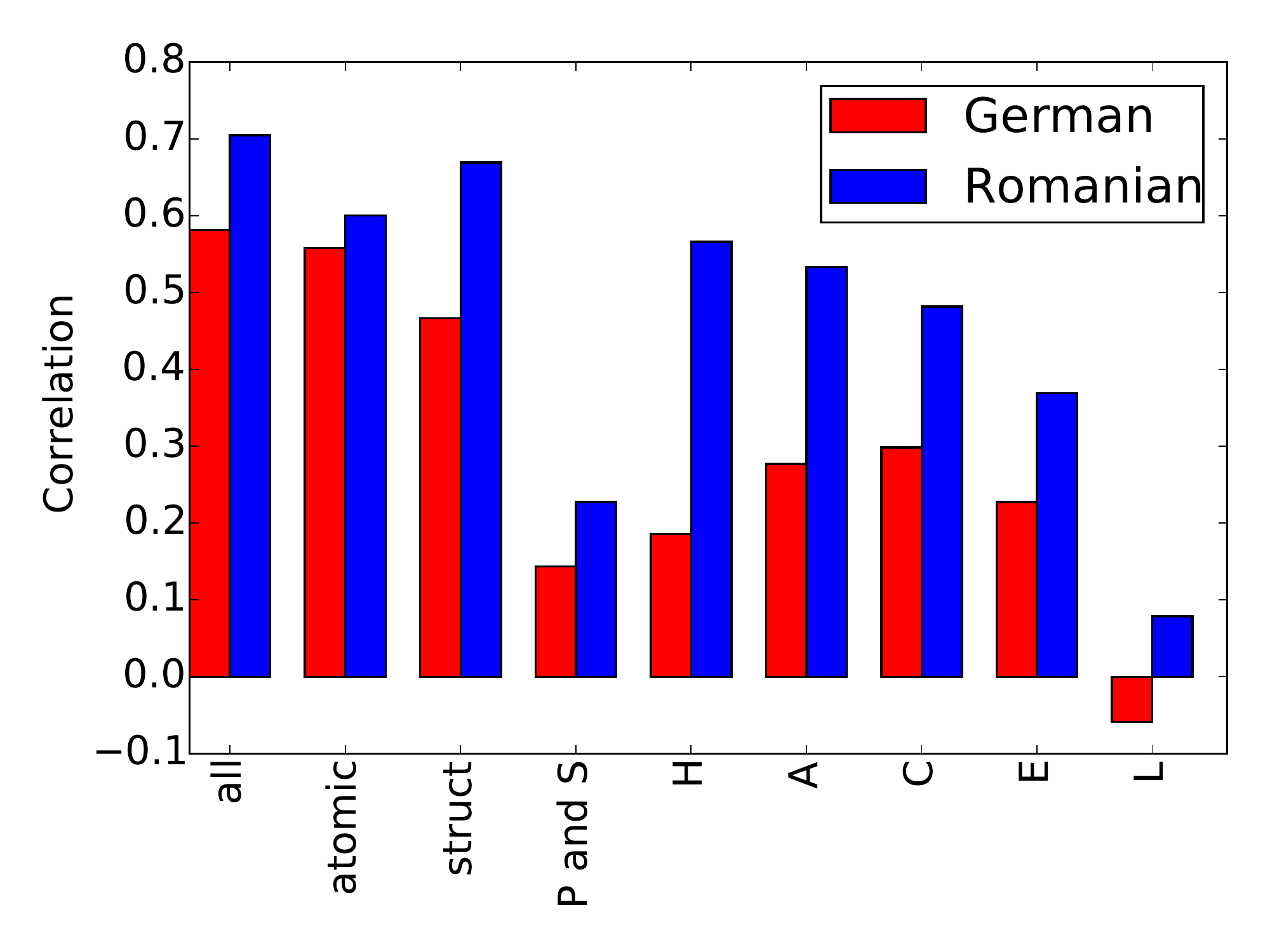}
\caption{Pearson correlation of HUME vs. DA scores for en-ro and en-de.
Each bar represents a correlation between DA and an aggregate HUME score
based on a sub-set of the units (\#nodes for the en-de/en-ro setting in brackets):
all units (``all'', 8624/10885), atomic (``atomic'', 5417/6888)
and structural units (``struct'', 3207/3997),
and units by UCCA categories: Scene main relations
(i.e, Process and State units; ``P and S'', 954/1178), Parallel Scenes (``H'', 656/784),
Participants (``A'', 1348/1746),
Centres (``C'', 1904/2474), elaborators (``E'', 1608/2031) and linkers (``L'', 261/315).
\label{fig:dacorrelationtypes}}
\end{figure}

Figure~\ref{fig:dacorrelationtypes} presents an analysis of HUME's correlations with DA by HUME unit type,
an analysis enabled by HUME's semantic decomposition. 
For both target languages, correlation is highest in the 'all' case, supporting our claim for the value of aggregating over a wide range of semantic phenomena. Some types of nodes predict the DA scores better than others. HUME scores on As correlate more strongly with DA than scores on Scene Main Relations (P+S). Center nodes (C) are also more correlated than elaborator nodes (E), which is expected given that Centers are defined to be more semantically dominant. Future work will construct an aggregate HUME score which weights the different node types according to their semantic prominence. 

\begin{figure*}[t]
\begin{tikzpicture}
\node(tree) [] {\includegraphics[width=1\textwidth]{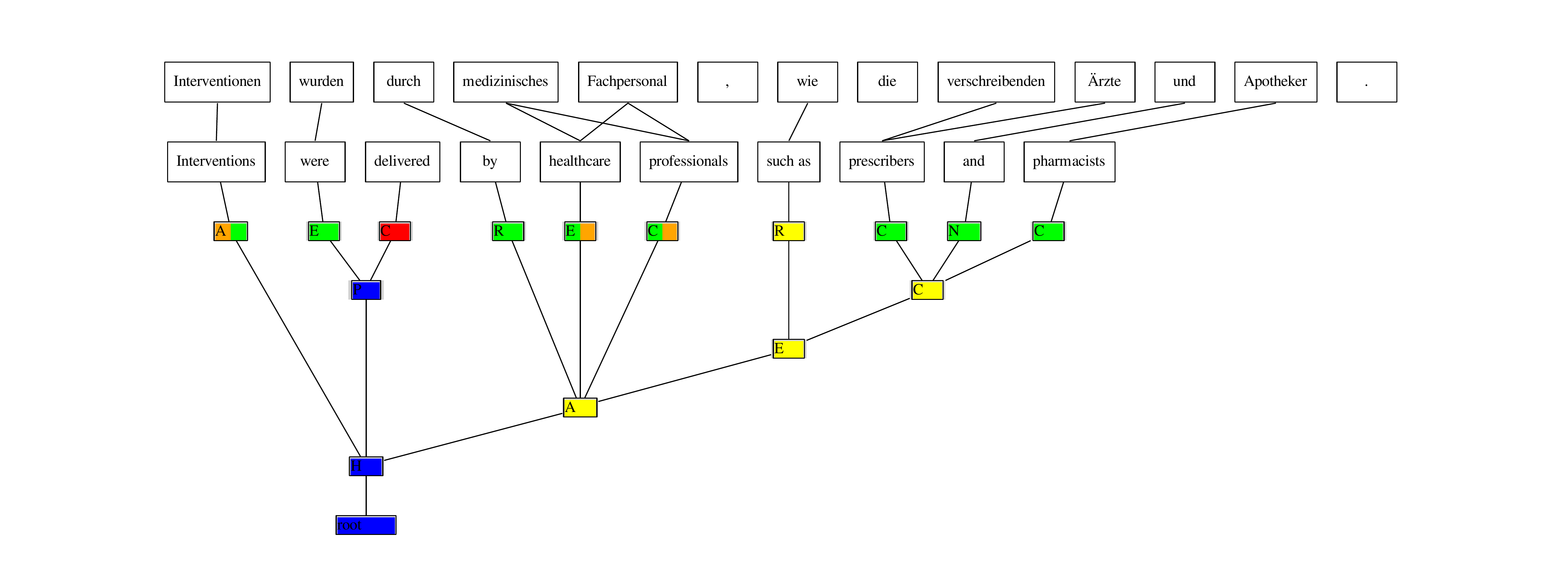}};
\node [below right = -2cm and  -8cm of tree]{\includegraphics[width=0.4\textwidth]{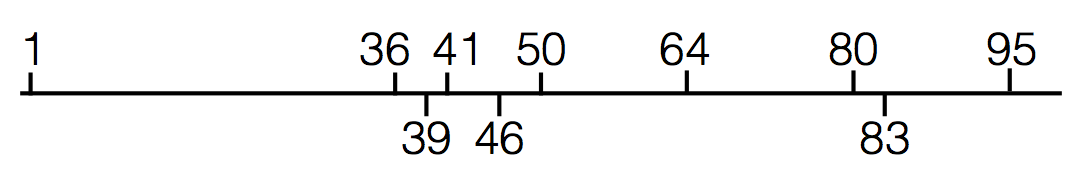}};
\end{tikzpicture}
\caption{An example of a doubly annotated HUME en-de sentence. The color of the nodes correspond to their
  HUME label: yellow (A), blue (B), and atomic nodes with their traffic light color. Nodes in which the annotators
  disagree are split in color. The scale at the bottom right shows a sample of 10 raw DA assessments
  for this example. }
\label{fig:humeeg}
\end{figure*}
%


Figure~\ref{fig:humeeg} presents an example of a doubly HUME-annotated English-German example
accompanied by 10 raw DA scores assigned to that sentence.
The figure illustrates the conceptual difference between the measures:
while DA standardises and averages scores across annotators to denoise the
crowd-sourced raw data, thus obtaining a single aggregate score,
HUME decomposes over a combinatorial structure, thus allowing to
localize the translation errors. We now turn to comparing HUME to a more
conceptually-related  measure, namely HMEANT.


\section{Comparison with HMEANT}\label{sec:hmeant_comp}

HMEANT is a human MT evaluation metric that measures the overlap between the translation a reference
in terms of their SRL annotations.
In this section we present a qualitative comparison between HUME and HMEANT,
using examples from our experimental data.


\paragraph{Verbal Structures Only?}

HMEANT focuses on verbal argument structures, ignoring other pervasive phenomena such as non-verbal predicates and inter-clausal relations. Consider the following example:

\begin{center}
\begin{tabular}{lp{5.4cm}}
Source & \small a coronary angioplasty may not be technically possible \\
Transl.& \small eine koronare Angioplastie kann nicht technisch m{\"o}glich \\
Gloss & \small a coronary angioplasty can not technically possible \\
\end{tabular}
\end{center}


The German translation is largely correct, except that the main verb ``sein'' (``be'') is omitted.
While this may be interpreted as a minor error, HMEANT will assign the
sentence a very low score, as it failed to translate the main verb.

It is also relatively common that verbal constructions are translated
as non-verbal ones or vice versa. Consider the following example:

\begin{center}
\begin{tabular}{lp{5.4cm}}
Source & \small ... tend to be higher in saturated fats \\
Transl.& \small ... in der Regel h{\"o}her in ges{\"a}ttigte Fette \\
Gloss & \small ... as a rule higher in saturated fats \\
\end{tabular}
\end{center}

The German translation is largely correct, despite the grammatical divergence,
namely that the English verb ``tend''
is translated into the German prepositional phrase ``in der Regel'' (``as a rule'').
HMEANT will consider the translation to be of poor quality as there is no German verb
to align with the English one.

\com{
In other cases, HMEANT's does not to penalize errors. For instance, as HMEANT is defined as an aggregate
of the scores given to the slot fillers, it cannot assign partial scores to the predicates themselves in
cases of a divergent tense or a negation flip. Consider the following:

\begin{center}
\begin{tabular}{lp{5.4cm}}
Source & \small ... don't skip meals ... \\
Transl.& \small ... nicht {\"u}berspringen Mahlzeiten ... \\
Gloss & \small ... don't skip meals ... \\
\end{tabular}
\end{center}

The translation is mostly understandable, although the verb ``{\"u}berspringen'' is incorrectly inflected.
HUME allows marking it as Orange (as it was  indeed annotated in our experiments), while
HMEANT allows either to mark it as correct, or otherwise mark the entire frame as a ``no match''.
}

We conducted an analysis of the English UCCA Wikipedia corpus (5324 sentences) in order to assess the
pervasiveness of three phenomena that are not well supported by HMEANT.\footnote{Argument structures and
linkers are explicitly marked in UCCA. Non-auxiliary instances of ``be'' and nouns are identified using
the NLTK standard tagger. Nominal argument structures are here Scenes whose Main Relation is headed by a noun.}
First, copula clauses are treated in HMEANT simply as instances of the main verb ``be'', which generally does not convey the meaning of these clauses. They appear in 21.7\% of the sentences, according to conservative estimates that only consider non-auxiliary instances of ``be''. Second, nominal argument structures, ignored by HMEANT, are in fact highly pervasive, appearing in 48.7\% of the sentences. Third, linkers that express inter-relations between clauses (mainly discourse markers and conjunctions) appear in 56\% of the sentences, but are again ignored by HMEANT. For instance, linkers are sometimes omitted in translation, but these omissions are not penalized by HMEANT. The following is such an example from our experimental dataset:

\begin{center}
\begin{tabular}{lp{5.4cm}}
Source & \small However, this review was restricted to ... \\
Transl.& \small Diese {\"U}berpr{\"u}fung bescr{\"a}nkte sich auf ... \\
Gloss & \small This review was restricted to ... \\
\end{tabular}
\end{center}


We note that some of these issues were already observed
in previous applications of HMEANT to languages other than
English. See \perscite{birch-EtAl:2013:WMT} for German, \perscite{bojar:wu:ssst:2012} for
Czech and \perscite{chuchunkov-tarelkin-galinskaya:2014:SSST-8} for Russian.


%
%
%


\paragraph{One Structure or Two.}
HUME only annotates the source, while HMEANT relies on
two independently constructed structural annotations, one for the reference and one
for the translation.
Not annotating the translation is appealing as it is often impossible to assign a
semantic structure to a low quality translation.
On the other hand, HUME may be artificially boosting the perceived understandability
of the translation by allowing access to the source.

\paragraph{Alignment.}
In HMEANT, the alignment between the reference and translation structures is a key
part of the manual annotation. If the alignment cannot be created, the 
translation is heavily penalized.
\perscite{bojar:wu:ssst:2012} and \perscite{chuchunkov-tarelkin-galinskaya:2014:SSST-8}
argue that the structures of the reference and of an accurate translation
may still diverge, for instance due to a different
interpretation of a PP-attachment, or the verb having an additional modifier in
one of the structures. It would be desirable to allow modifications to
the SRL annotations at the alignment stage,
to avoid unduly penalizing such spurious divergences.

The same issue is noted by \perscite{lo:wu:reliability:2014}:
the IAA on SRL dropped from 90\% to 61\% 
when the two aligned structures were from two different annotators.
HUME uses automatic (word-level) alignment, which only
serves as a cue for directing the attention of the annotators.
The user is expected to mentally correct the
alignment as needed, thus circumventing this difficulty.


\paragraph{Monolingual vs. Bilingual Evaluation.}
\label{src-vs-ref}
HUME diverges from HMEANT and from shallower measures
like BLEU, in not requiring a reference.
Instead, it directly compares the source and the translation.
This requires the employment of bilingual annotators, but has the benefit of avoiding
using a reference, which is never uniquely defined, and may thus 
lead to unjustly low scores where the translation is a paraphrase of the reference.
If only monolingual annotators are available, the HUME evaluation could be performed
with a reference sentence instead of with the source. This, however,
would risk inaccurate judgements due to the naturally occurring differences
between the source and its reference translations.

\section{Conclusion}\label{sec:conclusion}

We have introduced HUME, a human semantic MT evaluation measure which addresses
a wide range of semantic phenomena. We have shown that it can be reliably and 
efficiently annotated in multiple languages,
and that annotation quality is robust to sentence length.
Comparison to direct assessments further support HUME's validity.
We believe that HUME, and a future automated version of HUME,
allows for a finer-grained analysis of translation quality,
and will be useful in informing the development of a more semantically aware
approach to MT.

All annotation data gathered in this project, together with analysis scripts, is available
online\footnote{\url{https://github.com/bhaddow/hume-emnlp16}}.

\section*{Acknowledgments}

This project has received funding from the European Union's Horizon 2020 research and innovation
programme under grant agreement \GrantNo\ (\ProjectName).


\bibliography{main}
\bibliographystyle{emnlp2016}

\end{document}